\documentclass[runningheads]{llncs}
\usepackage{graphicx}
\usepackage{multirow}
\usepackage{caption}
\usepackage{subcaption}
\usepackage{booktabs}
\usepackage[sorting=none]{biblatex} 
\usepackage{array}
\usepackage{tabularx}
\usepackage{dcolumn}
\begin{document}
\title{Paperswithtopic: Topic Identification from Paper Title Only}
%
\author{Daehyun Cho\inst{1}\orcidID{0000-0003-2925-6459} \and \newline
Christian Wallraven\inst{2}\orcidID{0000-0002-2604-9115}}

\authorrunning{Cho and Wallraven}
%
\institute{Department of Artificial Intelligence, Korea University, Seoul, Korea\inst{1} \newline
\email{1phantasmas@korea.ac.kr}\newline
Department of Artificial Intelligence \& Department of Brain and Cognitive Engineering, Korea University, Seoul, Korea\inst{2}  \newline
\email{wallraven@korea.ac.kr}
}

\maketitle              
\begin{abstract}
The deep learning field is growing rapidly as witnessed by the exponential growth of papers submitted to journals, conferences, and pre-print servers. To cope with the sheer number of papers, several text mining tools from natural language processing (NLP) have been proposed that enable researchers to keep track of recent findings. 
In this context, our paper makes two main contributions: first, we collected and annotated a dataset of papers paired by title and sub-field from the field of artificial intelligence (AI), and, second, we present results on how to predict a paper's AI sub-field from a given paper title only. Importantly, for the latter, short-text classification task we compare several algorithms from conventional machine learning all the way up to recent, larger transformer architectures. Finally, for the transformer models, we also present gradient-based, attention visualizations to further explain the model's classification process. All code can be found online\footnote{Code available here: \url{https://github.com/1pha/paperswithtopic}}.

\keywords{Natural Language Processing \and Sequence Classification \and Deep Learning \and Model Comparison.}
\end{abstract}

\section{Introduction}
The field of artificial intelligence (AI) has undergone a rapid expansion over the past decades, fuelled especially by the advent of deep learning (DL) techniques in 2012 \cite{krizhevsky2012imagenet}. As exemplary statistics, around 370,000 papers from the United States and another 320,000 papers from China alone were published in the field of AI between 1997 and 2017 \cite{statistia}. Publication growth has been largely exponential over the past few years, which has increased the need for researchers and users in AI for text-mining tools that help to search, organize, and summarize the ``flood" of papers published every month. Popular bibliography tools have started to assist people through recommendation and ordering functions, including Mendeley \cite{zaugg2011mendeley} and EndNote \cite{hupe2019endnote}, for example. However, these applications currently still use ``standard" keyword association models \cite{mendeley}, and hence may have trouble in customizing recommendations and sorting according to specific user requirements. With the advent of more powerful natural language processing (NLP) tools over the past decade, additional methods for paper recommendation or summarization have been introduced, such as semantic-scholar \cite{cachola2020tldr}, which uses a series of NLP models to summarize a given paper into one sentence.   

In the present work, we aim to add to this field of recommendation and summarization systems by presenting a framework that is able to predict a paper's sub-field (in AI) from its title alone. One of the most popular resources for AI researchers is the pre-print service arXiv\footnote{https://arxiv.org/}, at which AI-related papers can be ``published" without peer-review. For predicting research fields, a dataset based on arXiv papers from various domains such as physics, computer-science, or chemistry \cite{clement2019arxiv} has already been created - this dataset, however, only contains these broad research domains, and does not indicate sub-fields, such as natural-language-processing or computer-vision for AI, for example. Such a more detailed classification, however, is crucial for a better organizing of papers - especially those that do not contain keywords, or non-curated, noisy keywords. 

In order to address this issue, here we first present a dataset that pairs current papers in the field of AI with their corresponding sub-field. We then show results from a series of algorithms that try to predict the sub-field from the paper's title - these algorithms range from conventional machine learning methods all the way up to recent, deep-learning-based transformer models, thus setting benchmarks for this task. Using methods from explainable AI, we also show how the latter, transformer-based models arrive at their decision. 

\section{Related Work}
\label{related}
The basic task for addressing the title-to-sub-field classification problem is called ``short sequence classification" - one of the most fundamental tasks in the NLP field \cite{goldberg2017neural}. In the following, we review methods for this task ranging from conventional machine learning models to current transformer-based architectures. 

\vspace{-0.3cm}
\subsection{Conventional Machine Learning Methods}
\vspace{-0.2cm}
Initial learning-based text classification methods first found wide adoption with the introduction of the \emph{Naive Bayes} \cite{lewis1998naive} model, still one of the most commonly-used spam mail classification methods \cite{metsis2006spam}. Although this model treats words as individually-independent, it still demonstrates high performance in short text classification tasks. However, without sophisticated preprocessing algorithms, the naive bayes classifier cannot handle multi-label tasks easily. 
The NLP field then introduced tree-based classifiers, such as \emph{decision trees} or ensemble versions as \emph{random forests} for such classification tasks - however, to sometimes mixed results that did not surpass existing machine learning models \cite{xu2012improved}. Since tree models are powerful at pruning tabular data, informative features with many different unique words may cause these models to still pay attention to uninformative features in order to keep the representation rich enough. Several approaches addressed this problem \cite{rf_semantic}, usually by boosting the ensembling of individual weak tree classifiers, such as \emph{XGBoost} \cite{chen2016xgboost}, or \emph{LightGBM} \cite{ke2017lightgbm}. 

\subsection{Recurrent Networks}
Among methods based on neural networks (NN), one of the earliest approaches to sequence classification is the \emph{recurrent neural network} \cite{rumelhart1985learning, osti_6910294} (RNN) framework, in which sequential data is fed into a neural network with the goal of predicting a future state or sequence. Given that text forms a natural sequence (of letters, words, and/or sentences), RNNs were quickly adopted for text classification \cite{guo2018cran} once large-enough corpora and powerful GPU processing became available. 
A critical issue with standard RNN models is that their gradients vanish in long sequences, which prompted the development of the \emph{long-short term memory} (LSTM) \cite{hochreiter1997long} architecture. Although the gates introduced in the LSTM framework managed to deal with the vanishing gradient problem, the framework overall suffered from exceedingly high computational cost due to the large amount of matrix multiplications per sequence. To reduce these computational costs, derivatives of the LSTM model, such as \emph{gated recurrent unit} (GRU) \cite{DBLP:journals/corr/ChoMGBSB14} were introduced. Overall, recurrent models already increased accuracy by a significant margin compared to conventional machine learning algorithms. 

\subsection{Attention-based Models}
A further development in sequence processing and in NLP in general came from the development of the \emph{attention mechanism}, which was first introduced into one of the recurrent network models \emph{seq2seq} \cite{DBLP:journals/corr/BahdanauCB14} - a model that is still widely used in machine translation tasks in NLP. The main idea of this mechanism is to find the relationship between an input sequence with a hidden vector. Through this attentional connection, models became better able at capturing relations between sequences at even longer distances and better fidelity.

Perhaps one of the most significant developments in NLP in recent years was made by the introduction of the \emph{transformer} architecture - a further development of the attention mechanism \cite{DBLP:journals/corr/VaswaniSPUJGKP17}. This model and its subsequent derivatives have become the de-facto benchmark for nearly all language-related tasks over the past years. Derivatives of the original transformer architecture include Google BERT \cite{DBLP:journals/corr/abs-1810-04805}, Open-AI GPT \cite{DBLP:journals/corr/abs-2005-14165} (now in its third, to date largest iteration), and ELECTRA \cite{DBLP:journals/corr/abs-2003-10555}. Similarly to the early issues of LSTM, however, the attention mechanism inside the transformer architecture has in principle quadratic sequence length complexity - an issue that was addressed in updated architectures, such as Performer \cite{DBLP:journals/corr/abs-2009-14794} based on a kernel method, for example. Similarly, ALBERT \cite{DBLP:journals/corr/abs-1909-11942} tried to reduce the computational cost by using extensive weight sharing and factorized embedding parametrization. 

\subsection{Recommendation systems for paper management}
As our application domain is similar to recommendation systems, we will briefly review some relevant methods for these systems as well: these often use methods based on collaborative filtering, sometimes adding content-based filtering or graph-based methods \cite{bai2019scientific}. The core idea here is to translate the interaction between user and item into a pivot table matrix. Similarities between users or items based on this matrix are then found by using cosine similarity or matrix factorization methods to derive embedding vectors. A common issue with these approaches is the ``cold-start problem'': new users or items lack interaction with others, which makes it hard to retrieve their embedding vectors. Deep learning methods that are not based on user information may be able to find such representations better. Services that are make use of such methods are Mendeley \cite{mendeley}, arxiv-sanity, or semantic-scholar - the latter being based on the recent, state-of-the-art GPT-3 transformer model \cite{cachola2020tldr, DBLP:journals/corr/abs-2005-14165}.

A scientific dataset that was created for a short sentence classification task similar to the one we deal with, is PubMed 200k \cite{DBLP:journals/corr/abs-1710-06071}. This is a collection of sentences labeled according to the section they belong to, such as `Results' or `Methods', and has been used with different NLP models - it does not, however, give any insights about the specific paper topic they belong to.

\section{Methods}
\label{methods}
The bulk of the experimental work was implemented with Python and the PyTorch deep learning framework \cite{NEURIPS2019_9015}. Most of the conventional machine learning models were implemented through scikit-learn \cite{scikit-learn}, with recent transformer models using the huggingface library \cite{DBLP:journals/corr/abs-1910-03771}. Experiment maintenance and logging made use of wandb.ai \cite{wandb}\footnote{wandb project link: \url{https://wandb.ai/1pha/paperswithtopic}}.

In the following, we describe the overall pipeline from data collection, pre-processing, model implementation, hyper-parameter optimization, to result collection. Overall classification performance was measured using the area under the receiver operating curve (AUROC, \cite{hanley1982meaning}). 

\subsection{Dataset}
One of the most popular resources for AI researchers is paperswithcode\footnote{paperswithcode: \url{https://paperswithcode.com/}}. This site allows researchers to see state-of-the-art models based on different tasks. From their publicly-available database, we collected the meta-data resulting in 49,980 available papers. The full meta-data, including abstracts or publication date, was saved in \verb+.json+ format, and the paper title paired with corresponding sub-field category labels were saved in \verb+.csv+ format in a pivot table scheme. There were a total of sixteen  imbalanced labels: adversarial, audio, computer-code, computer-vision, graphs, knowledge-base, medical, methodology, miscellaneous, music, natural-language-processing, playing-games, reasoning, robots, and speech. Since the label methodology introduced a large amount of skew into the distribution, we excluded this label from the training phase, resulting in a total of 15 labels. From this dataset, we split stratified train, validation and test sets with a 90/5/5 ratio, resulting in 34,572 papers for training and 1,921 papers for validation and test set, respectively. 

\subsection{Preprocessing}
Preprocessing the input into a correct, machine-learning suitable format is one of the most important parts of any NLP pipeline. Typical boilerplate pipelines include removing extra whitespaces and special characters, as well as lowercasing, which we followed in this order. Pipelines also often include removing numbers or expanding contractions (i.e. \verb+I'll+ to \verb+I will+) - here, we want to keep the numbers as they may be important predicting factors (e.g., GPT-``3"); in contrast, contractions rarely happen in paper titles, which is why we skipped these two preprocessing stages. Moreover, lemmatization is often done to improve generalization, but again, titles do not tend to include much tense shifting, hence this step was also not included. Finally, the pre-processed titles were tokenized into an index and filled by \verb+<pad>+ for sequences shorter than the designated maximum sequence length (in our case set to 64 characters; sequences longer than the 64 characters would get truncated, but the paperswithcode dataset did not contain any title longer than 50 characters). For deep learning models, these sequences are usually further embedded into vectors, in order to represent words with contextual information. This procedure can be done via embedding matrices included inside the deep learning methods, or via libraries that assist with this process such as FastText \cite{DBLP:journals/corr/JoulinGBDJM16} or Word2Vec \cite{DBLP:journals/corr/abs-1301-3781} - these approaches will be discussed more in Section \ref{rnn_train}.

\subsection{Conventional Machine Learning Models}
We implemented the following conventional machine learning models: Gaussian naive Bayes, complement naive Bayes, extra tree classifier, random forest classifier, k-nearest neighbor classifier, AdaBoost classifier, LightGBM, and XGBoost. Hyper-parameters for all models were used with their default values as implemented in scikit-learn.

Since our task is a multi-label classification task with some instances having multiple labels, which some methods cannot handle properly, models were fitted with each label sequentially, hence treating labels as individually independently distributed.

\subsection{Recurrent Neural Network Models}
For standard RNN models, we used a ``vanilla" RNN \cite{rumelhart1985learning}, as well as LSTM \cite{hochreiter1997long} and its derivate GRU \cite{DBLP:journals/corr/ChoMGBSB14}. All RNNs used binary cross-entropy loss for each label and then used a mean loss across all labels for each batch. We employed the Adam optimizer with a learning rate of $10^{-4}$ and default update parameters of $\beta_1=0.9$ and $\beta_2=0.999$. Early stopping was used on the validation set with a patience of 15 epochs.

Variations to these RNN models \cite{jozefowicz2015empirical} can be done by stacking more layers into the network or by increasing the embedding dimensions, which may affect performance. We therefore performed an additional grid search on these two parameters. 

\vspace{-0.2cm}
\subsection{Transformer Models}
\vspace{-0.2cm}
A number of high-performance transformer variations came out starting with BERT \cite{DBLP:journals/corr/abs-1810-04805}. Here, we compared BERT \cite{DBLP:journals/corr/abs-1810-04805}, ELECTRA models \cite{DBLP:journals/corr/abs-2003-10555} starting with default parameters of a single layer of 8 attention heads, and a hidden dimension of 128. Furthermore, for the ELECTRA model we performed a grid search on modulating the number of layers and number of heads.

\vspace{-0.2cm}
\subsection{Word embeddings}
\vspace{-0.2cm}
For both the RNNs and the Transformer models, we also tested three different ways to embed/process the tokenized inputs: first, we created a trainable lookup table for each token, such that the matrix has its shape of ($\#$ of tokens, $\#$ of dimensions). This is an end-to-end process in that a single model learns embeddings and classification. For our second and third method, we used the two embedding methods of word2vec and FastText. These methods learn to embed the tokens into a vector first, with the neural networks getting trained on the classification task on these vectors. This can be regarded as two-stage model with the optimizer, however, not being able to affect the embedding itself.

\vspace{-0.2cm}
\subsection{Pooling method}
\vspace{-0.2cm}
For transformer methods, researchers typically use the last hidden state of the transformer outputs to classify sequences (last token pooling in Table \ref{table:dl}). The Huggingface API, however, also supports sequence classification models for each transformer architecture that can use the \emph{first} hidden states for classification. For this, the first hidden state is internally pooled as a process of point-wise feedforward action - similar to the attention calculation proposed in \cite{DBLP:journals/corr/VaswaniSPUJGKP17}. For the BERT and ELECTRA models, we therefore also compared the first and last token pooling classification results.

\vspace{-0.2cm}
\subsection{Visualization}
\vspace{-0.2cm}
It is possible for transformer models to retrieve an ``attention score" that illustrates the relation between sequences. The standard Bahdanau attention mechanism \cite{DBLP:journals/corr/VaswaniSPUJGKP17} yields an attention matrix consisting of a set of scores for each sequence (normalized to 1). This matrix encodes which relations between sequences affect the final output as they receive point-wise feedforward from the final output. This method works well for seq2seq models \cite{DBLP:journals/corr/BahdanauCB14}, but not directly for transformer architectures, as these have multiple layers each with multiple heads. Preliminary research \cite{DBLP:journals/corr/abs-1908-04211, DBLP:journals/corr/abs-1902-10186} has shown that a resulting attention matrix in the latest transformer architectures only indirectly visualizes their predictions (it can still be used, however, to follow the training process  \cite{DBLP:journals/corr/abs-1908-04626, DBLP:journals/corr/abs-1906-04341}).

Here, we tried to find a visualization of the models based on the idea of Gradient-weighted Class Activation Mapping (Grad-CAM) \cite{selvaraju2017grad} instead, which uses the gradient to locate inputs (originally pixels) important for the models' decision. The main logic here is to retrieve the gradient of the target layer during optimization, convert it to a mean averaged kernel and to overlay this kernel with the original image to highlight the ``important" input. For this, we retrieved the gradients that were backward-passed from the embedding matrix to the tokenized input in our transformer architectures (using the trainable lookup embedding), and normalized the result using the $L2$-norm for each sequence to get an attention/importance score.

\vspace{-0.3cm}
\section{Results}
\vspace{-0.4cm}
\subsection{Conventional Machine Learning Models}
Table \ref{table:ml} shows the AUROC for the conventional machine learning models. Extra Tree and K-Nearest Neighbour had low performance close to chance (AUROC = 0.5 equals a random guess). It is especially interesting in this context that both naive Bayes models did not result in significant performance increases given their general use as spam-filters. We assume that these models are more vulnerable to unseen words (e.g. ``flickering" which did not appear in the training set) due to their naive multiplication of word probabilities. Beyond the Extra Tree model, other tree-based models, however, showed higher performance. Nonetheless, these performance levels seem not sufficient enough for use in practical applications.

\begin{table}[h!]
  \begin{center}
  \begin{tabular}{p{0.25\textwidth}p{0.35\textwidth}c}
    \toprule
    Types    & Models           & AUROC \\
    \midrule
    Machine Learning & Extra Tree             & 52.4 $\pm$ 0.02\\ 
                     & Complement Naive Bayes & 52.5 $\pm$ 0.12\\
                     & Naive Bayes            & 52.9 $\pm$ 0.05\\
                     & K-Nearest Neighbour    & 53.2 $\pm$ 0.03\\
                     & Random Forest          & 56.4 $\pm$ 0.05\\
                     & AdaBoost               & 59.8 $\pm$ 0.06\\
                     & LightGBM               & 63.9 $\pm$ 0.08\\
                     & XGBoost                & 66.8 $\pm$ 0.09\\ 
    \bottomrule
  \end{tabular}
\end{center}
\caption{AUROC comparison between machine learning models, using tokenized input. Error indicates Standard Error of the Mean (SEM) across 10 folds.}
\label{table:ml}
\vspace{-0.8cm}
\end{table}

\vspace{-0.4cm}
\subsection{Recurrent Neural Network Models} \label{rnn_train}
Using a single layer LSTM and GRU models already outperformed all conventional methods (cf. first column in Table \ref{table:rnn} versus Table \ref{table:ml}). Improvements across number of layers or architectures, however, was less visible. Adding more layers to the ``vanilla" RNN resulted in a clear decrease in performance for all values of the hidden dimension, most likely due to the issue of vanishing gradients which resulted in unstable optimization. For the other two RNNs, however, stacking more layers either kept performance or resulted in lower overall performance. Best performance for both LSTM and GRU models was obtained with the maximum amount of hidden dimensions and single layer. Performance for the GRU model seemed to be more stable as the number of layer increased compared to the LSTM model that decreased performance with increasing number of layers.

\begin{table}[h!]
  \begin{center}
  \begin{tabular}{p{0.2\textwidth}p{0.1\textwidth}<{\centering}p{0.1\textwidth}<{\centering}p{0.1\textwidth}<{\centering}p{0.1\textwidth}<{\centering}p{0.1\textwidth}<{\centering}}
    \toprule
    Model & \#H    & \multicolumn{4}{c}{\#L}     \\
    \cmidrule(r){3-6}
        &     &  1 & 2    & 3    & 4  \\
    
    \midrule
    RNN     & 128   & 61.7  & 56.0  & 59.7 & 55.3 \\
            & 256   & 54.1  & \textbf{62.8}  & 54.1 & 49.8 \\
            & 512   & 52.8  & 51.0  & 50.7 & 50.4 \\
                    
    \midrule
    LSTM    & 128   & 85.9  & 85.1  & 83.3 & 82.9 \\
            & 256   & 86.8  & 85.7  & 85.2 & 85.4 \\
            & 512   & \textbf{90.4}  & 89.4  & 87.5 & 75.8 \\
                
    \midrule
    GRU     & 128   & 85.4  & 87.1  & 88.0 & 87.7 \\
            & 256   & 90.2  & 88.8  & 89.0 & 88.2 \\
            & 512   & \textbf{92.2}  & 88.9  & 89.2 & 88.0 \\
                     
    \bottomrule
  \end{tabular}
  
  \end{center}
  \caption{Grid search result for hyper-parameters for recurrent neural networks (AUROC): \#H=hidden size; \#L=the number of layers. Bolded values indicate best performance for each model.}
  \label{table:rnn}
  \vspace{-0.8cm}
\end{table}

\vspace{-0.4cm}
\subsection{Transformer Models}
Table \ref{table:bert} shows the results of the grid search for the ELECTRA model that predicts with the result from first token pooling. Performance in all cases is high around 0.9 (except for one optimization failure shown in italics in Table \ref{table:bert}). The number of attention heads has limited influence across all parameters in most cases. However, the model fails to train with more layers with small hidden dimension even with large number of multiple heads. The size of the hidden state, however, results in a slight improvement across architectures and other hyper-parameters, with the best results reaching around 93.7 for the medium hidden dimensionality.

\begin{table}[htb!]
  \begin{center}
  \begin{tabular}{p{0.2\textwidth}p{0.1\textwidth}<{\centering}p{0.1\textwidth}<{\centering}p{0.1\textwidth}<{\centering}p{0.1\textwidth}<{\centering}p{0.1\textwidth}<{\centering}}
    \toprule
    Model & \#L   & \#A  & \multicolumn{3}{c}{\#H} \\
    \cmidrule(r){4-6}
                &           && 128    & 256    & 512  \\
    
    \midrule
    ELECTRA & 1           & 8         & 93.0  & 93.1  & 92.7 \\
    &            & 16        & 93.1  & 93.1  & \textbf{93.6} \\    
    &            & 32        & 93.6  & 93.4  & 93.6 \\
                
    \cmidrule(r){2-6}
    &2           & 8         & 91.5  & 92.9  & 88.4 \\
    &            & 16        & 92.5  & 93.7  & 90.8 \\
    &            & 32        & 92.6  & \textbf{93.8}  & 92.6 \\  
                     
    \cmidrule(r){2-6}
    &3           & 8         & \textit{53.7}  & 93.2  & 81.3 \\
    &            & 16        & \textit{53.1}  & \textbf{93.7}  & 87.5 \\
    &            & 32        & 91.2  & 93.1  & 92.2 \\  
                     
    \cmidrule(r){2-6}
    &4           & 8         & \textit{52.6}  & 93.5  & 74.4 \\
    &            & 16        & \textit{50.3}  & 93.3  & 84.4 \\
    &            & 32        & \textit{53.7}  & \textbf{93.7}  & 89.9 \\
                     
    \bottomrule
  \end{tabular}
  \end{center}
    \caption{Grid search result over number of layers, heads and hidden dimension for the ELECTRA Transformer Model pooled with first sequence label: \#L=the number of layers; \#A=the number of attention heads; \#H=hidden size. }
  \label{table:bert}
  \vspace{-0.8cm}
\end{table}

\subsection{Comparison of deep learning models; effects of embedding and token pooling}

Table \ref{table:dl} shows the AUROC results for the different RNNs, Transformer Models (for the two different pooling methods), and the effect of the tokenization embedding method. To make recurrent network models comparable to transformer models, we used 2 layers for recurrent network models while transformer models used a single layer with 8 attention heads. In the recurrent network and last token pooling transformer models, word2vec and FastText method surpassed the lookup matrix in all number of hidden dimensions, while first token pooling model showed high performance when lookup table matrix was used. Increasing the embedding dimension enhanced the result in most recurrent networks. On the other hand, transformer models with last token pooling often failed to optimize as the embedding size increased. There was also a performance drop with the increment of embedding dimension in first token pooling but not as much as last token pooling. Comparing this to the base BERT \cite{DBLP:journals/corr/abs-1810-04805} model that uses embedding dimension of 768 and shows high performance in general NLP tasks, we suppose that these effects are due to the small size of vocabulary and input of relatively short sequences in our application domain. We could also observe that first token pooling, in overall, slightly outperforms last token pooling. 

\begin{table}[htb!]
  \begin{center}
    \begin{tabular}{p{0.3\textwidth}p{0.15\textwidth}p{0.1\textwidth}ccc}
    \toprule
    Types    & Models           & \#E &\multicolumn{3}{c}{AUROC} \\
    & & &\ Lookup\  & \ Word2Vec\  & \ FastText\  \\
    \midrule
    Recurrent Networks  & RNN  & 128 & 55.1 & 60.7 & 55.9  \\
                        &      & 256 & 59.3 & 55.5 & 53.4  \\
                        &      & 512 & 65.7 & \textbf{66.9} & 62.9  \\
                        \cmidrule(r){2-6}
                        & LSTM & 128 & 79.1 & 83.8 & \textbf{89.7} \\  
                        &      & 256 & 88.5 & 89.9 & 89.3 \\
                        &      & 512 & 86.0 & 89.0 & 87.1  \\
                        \cmidrule(r){2-6}
                        & GRU  & 128 & 88.3 & 88.0 & 77.5 \\
                        &      & 256 & 88.2 & 88.4 & 74.0 \\
                        &      & 512 & 86.6 & 89.1 & \textbf{88.6}  \\
                     
    \midrule
    Last Token Pooling  & BERT    & 128 & 88.9 & 92.3 & 91.8\\
                        &      & 256 & 82.8 & 90.2 & \textbf{92.0}  \\
                        &      & 512 & \textit{56.7} & 80.7 & \textit{54.3}  \\
                        \cmidrule(r){2-6}
                        & ELECTRA & 128 & 91.3  & 92.1 & \textbf{92.4} \\
                        &      & 256 & 90.8 & 92.3 & 91.9  \\
                        &      & 512 & \textit{50.0} & \textit{50.0} & \textit{50.0}  \\
                     
    \midrule
    First Token Pooling  & BERT    & 128 & \textbf{92.8}  & 91.8 & 91.2 \\
                        &      & 256 & 91.9 & 91.8 & 91.2  \\
                        &      & 512 & 91.6 & 89.9 & 90.0  \\
                        \cmidrule(r){2-6}
                        & ELECTRA & 128 & \textbf{93.0} & 91.7 & 91.7\\
                        &      & 256 & 92.6 & 92.3 & 91.8  \\
                        &      & 512 & 92.7 & 89.9 & 89.6  \\
                     
    \bottomrule
  \end{tabular}
  \end{center}
\caption{AUROC comparison between deep learning models, using different embedding methods and embedding dimensionalities as well as different token pooling methods. Last and first token pooling refers to which token was pooled to determine the logits. Best results for each method are bolded, failed optimizations are shown in italics: \#E=embedding size.}
\label{table:dl}
\vspace{-0.8cm}
\end{table}

\subsection{Visualization}
\begin{figure}[h!]
     \centering
     \begin{subfigure}[b]{\textwidth}
         \centering
         \includegraphics[width=0.8\textwidth]{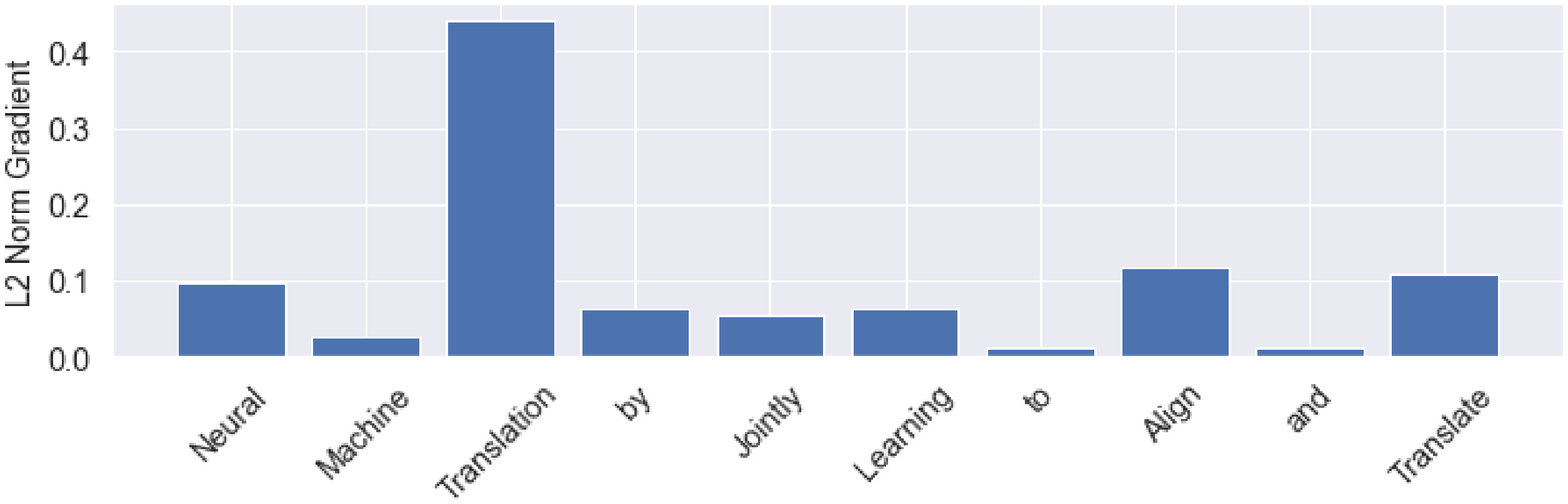}
         \caption{Model top-3 predictions: natural-language-processing, medical, speech. The top-1 prediction and true paper topic align.}
         \label{fig:seq2seq}
     \end{subfigure}
     
     \begin{subfigure}[b]{\textwidth}
         \centering
         \includegraphics[width=0.8\textwidth]{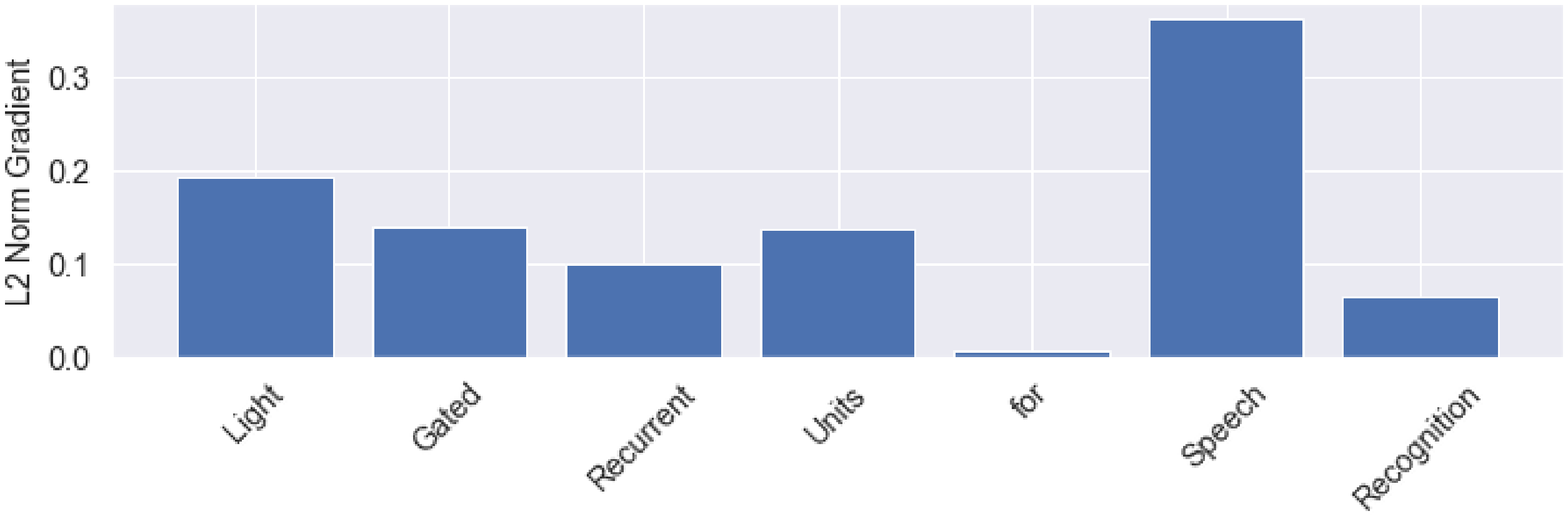}
         \caption{Model top-3 predictions: speech, natural-language-processing, time-series. Correct top-1 prediction, but note that the following predictions fit as well.}
         \label{fig:lightrnn}
     \end{subfigure}
     
     \begin{subfigure}[b]{\textwidth}
         \centering
         \includegraphics[width=0.8\textwidth]{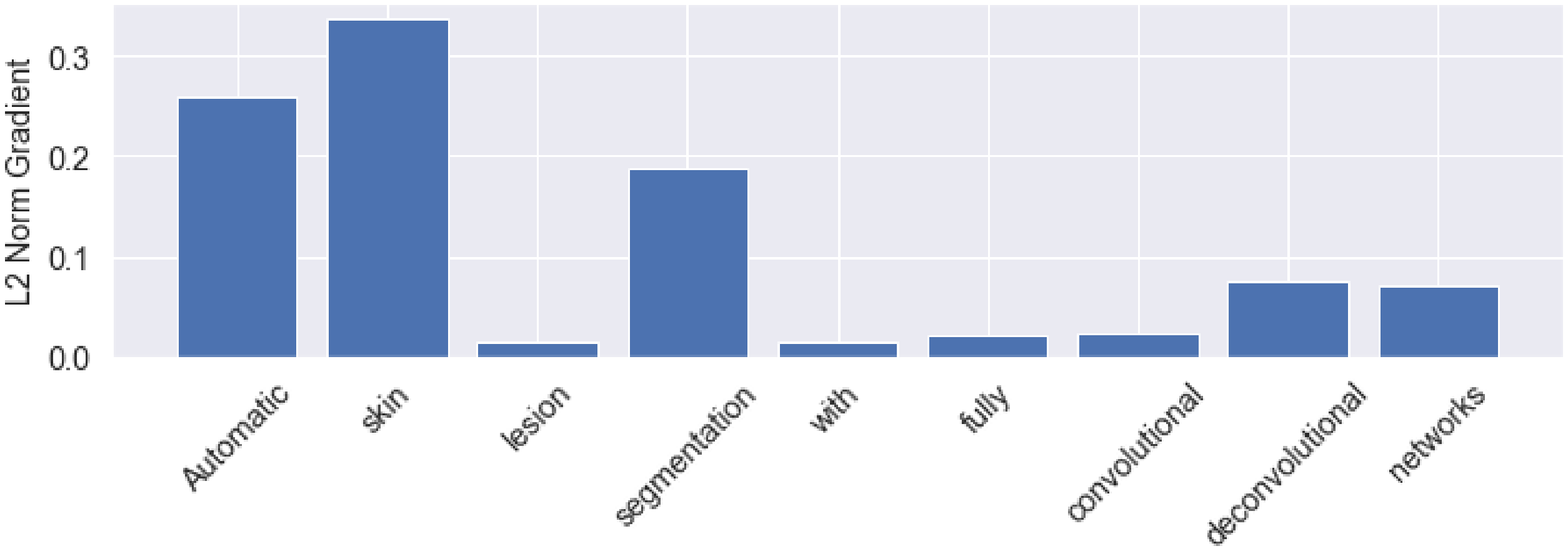}
         \caption{Model top-3 predictions: medical, computer-vision, robots. Correct for top-1 and top-2 predictions.}
         \label{fig:skinlesion}
     \end{subfigure}
     
     \begin{subfigure}[b]{\textwidth}
         \centering
         \includegraphics[width=0.8\textwidth]{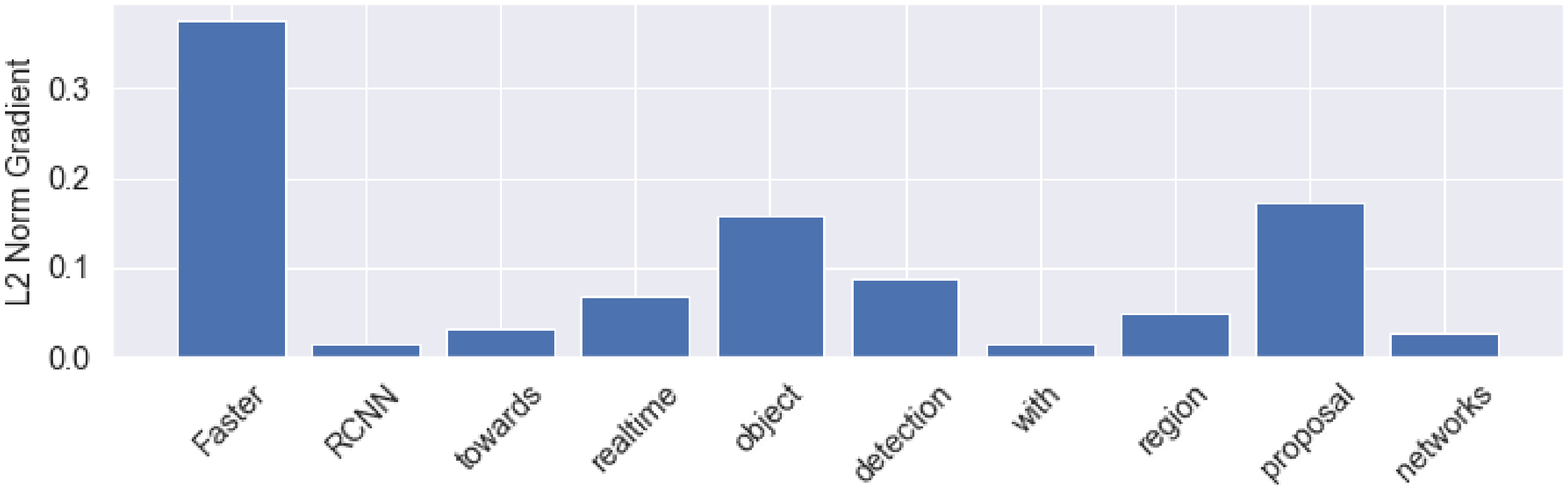}
         \caption{Model top-3 predictions: robots, computer-vision, medical. Top-1 prediction is not correct, top-2 is.}
         \label{fig:rcnn}
     \end{subfigure}

 \caption{Gradient-based Visualization of the output}
\label{fig:2}
\vspace{-0.8cm}
\end{figure}

 Results of the gradient-based visualization are demonstrated in Figures \ref{fig:2} for four different examples. 
 
 The first figure is the result for the paper title of ``Neural Machine Translation by Jointly Learning to Align and Translate'', which our model correctly predicts as natural-language-processing. Figure \ref{fig:seq2seq} clearly highlights ``translation" as the most important word for that decision. Other words (including prepositions or words that do not directly correlate with the prediction - at least from human intuition) have comparatively small values. Figure \ref{fig:lightrnn} and Figure \ref{fig:skinlesion} depict additional, successful cases in which the attention seems to rest on well-matching words.
 
 In contrast, Figure \ref{fig:rcnn} illustrates a failure case for the first category prediction for a paper with the title of ``Faster R-CNN towards real-time object detection with region proposal networks"--- this was mis-classified as belonging to the robots category. Here, the gradient-based visualization highlights the word ``Faster", as well as ``object'', which may have given rise to that wrong classification. Interestingly, the (correct) category of computer-vision, was the next-highest choice for the model, indicating that a Top-N accuracy evaluation method might result in even better scores for the transformer models.

\section{Discussion and Conclusion}
\label{discussion}
The main contribution of our work presented here was to compare different NLP models for a short-sequence-classification task in which a paper topic was to be predicted from its title. Based on our initial dataset on AI-topics, we found that current deep learning models fared best with a AUROC of 93.8 on the held-out test set, indicating excellent performance for this task domain. Less deep RNN models still achieved passable performance with a GRU, RNN reaching up to 92.2 with a large number of hidden states, yet at only one layer. Conventional machine learning models failed to reach satisfactory results indicating the superiority of sequence-based models in current NLP tasks even for---relatively---small datasets.

We also performed grid-search on a number of hyper-parameters of NN models and compared different tokenization and pooling methods for pre-processing and classification. Here, results showed that the number of hidden dimensions and also the embedding dimensionality need to be matched to the size of the dataset (in our case, towards the smaller end). Further experiments will need to be done with larger datasets to trace these performance patterns as the sample size increases.

In the future, we would like to extend our approach also to a wider range of scientific topics, expanding our dataset further. Similarly, additional features could be taken into account, such as authors, conferences, abstracts, keywords, or arXiv URLs. From a methodological point of view, we need to compare our results also to the popular topic modeling methods, such as Latent Dirichlet Allocation (LDA)  \cite{jelodar2019latent} or Latent Semantic Analysis (LSA) \cite{dumais2004latent}, which would make our analysis more complete. Additional extensions from the NN perspective concern methods that explicitly use weakly-labeled data (\cite{clement2019arxiv} such as present in the categories of arXiv) or self-supervised improvements.

\section{Acknowledgments}
This work was partly supported by Institute of Information \& Communications Technology Planning \& Evaluation (IITP) grants funded by the Korean government (MSIT) (No. 2019-0-00079, Department of Artificial Intelligence, Korea University; No.2021-0-02068-001, Artificial Intelligence Innovation Hub)

\AtNextBibliography{\small}
\printbibliography 
\end{document}